\documentclass[letterpaper]{article}
\usepackage{aaai2026}
\usepackage{times}
\usepackage{helvet}
\usepackage{courier}
\usepackage[hyphens]{url}
\usepackage{graphicx}
\usepackage{natbib}
\usepackage{caption}
\usepackage{algorithm}
\usepackage{algpseudocode}
\usepackage{amsthm}
\usepackage{amsmath}
\usepackage{xcolor}
\usepackage{amssymb}
\usepackage{mathtools}
\usepackage{epstopdf}
\usepackage{adjustbox}
\usepackage{subcaption}
\usepackage{bm}
\usepackage{booktabs}

\usepackage{newfloat}
\usepackage{listings}
\DeclareCaptionStyle{ruled}{labelfont=normalfont,labelsep=colon,strut=off} 
\floatstyle{ruled}
\newfloat{listing}{tb}{lst}{}
\floatname{listing}{Listing}
\title{Evaluating Knowledge Graph Complexity via Semantic, Spectral, and Structural Metrics for Link Prediction}
\author{
    Haji Gul\textsuperscript{\rm 1},
    Abul Ghani Naim\textsuperscript{\rm 1},
    Ajaz Ahmad Bhat\textsuperscript{\rm 1 $^*$}
}
\affiliations{
    \textsuperscript{\rm 1}School of Digital Science, Universiti Brunei Darussalam\\
    {(23h1710, ghani.naim, ajaz.bhat$^*$)@ubd.edu.bn}
}

\usepackage{bibentry}

\begin{document}

\maketitle

\begin{abstract}
Understanding dataset complexity is fundamental to evaluating and comparing link prediction models on knowledge graphs (KGs). While the Cumulative Spectral Gradient (CSG) metric \cite{branchaud2019spectral} —derived from probabilistic divergence between classes within a spectral clustering framework— has been proposed as a classifier-agnostic complexity metric—purportedly scaling with class cardinality and correlating with downstream performance, it has not been evaluated in KG settings so far.
In this work, we critically examine CSG in the context of multi-relational link prediction, incorporating semantic representations via transformer-derived embeddings. Contrary to prior claims, we find that CSG is highly sensitive to parametrisation and does not robustly scale with the number of classes. Moreover, it exhibits weak or inconsistent correlation with standard performance metrics such as Mean Reciprocal Rank (MRR) and Hit@1.
To deepen the analysis, we introduce and benchmark a set of structural and semantic KG complexity metrics. Our findings reveal that global and local relational ambiguity—captured via Relation Entropy, Node-level Maximum Relation Diversity, and Relation Type Cardinality—exhibit strong inverse correlations with MRR and Hit@1, suggesting these as more faithful indicators of task difficulty. Conversely, graph connectivity measures such as Average Degree, Degree Entropy, PageRank, and Eigenvector Centrality correlate positively with Hit@10.
Our results demonstrate that CSG’s purported stability and generalization‐predictive power fail to hold in link‐prediction settings, and underscore the need for more stable, interpretable, and task-aligned measures of dataset complexity in knowledge-driven learning.

\end{abstract}


\section{Introduction}
Knowledge graphs (KGs) underpin a wide array of high-impact applications, from recommendation systems~\cite{spillo2024evaluating} and question answering~\cite{zeng2025kosel} to drug discovery~\cite{zhang2025comprehensive,gul2025mucos}. By encoding facts as triples $(h,r,t)$, KGs support essential inference tasks such as link prediction $(h,?,t)$ and entity prediction $(h,r,?)$, which assess a model’s ability to recover missing relations or entities~\cite{gul2024contextualized,gul2025muco}. Despite steady improvements in knowledge graph embedding (KGE) techniques, benchmark performance—measured by Mean Reciprocal Rank (MRR) and Hits@k—varies dramatically across datasets. For example, models may achieve high MRR on CoDEx-S yet struggle on FB15k-237 or WN18RR, reflecting disparities that are poorly understood and complicating both model development and dataset selection.

Existing evaluation metrics quantify {\em how well} models perform, but they do not explain {\em why} a dataset is inherently difficult. Is low performance symptomatic of model limitations or indicative of high dataset complexity? Without principled measures of KG complexity, it is impossible to anticipate generalization performance, compare methods fairly, or guide dataset curation—e.g., to determine whether predicting rare drug–target pairs is intrinsically harder than recovering common relations. A robust complexity metric would (i) {\em quantify} dataset hardness across link‐prediction tasks, (ii) {\em predict} generalization before expensive downstream evaluations, and (iii) {\em unify} performance comparisons across heterogeneous benchmarks.

A number of prior works have explored facets of KG complexity, but none have provided a unified, task‐agnostic framework tailored to link‐prediction. Information‐theoretic approaches—such as Gaussian embeddings with differential entropy to capture semantic uncertainty~\cite{vilnis2015word}—demonstrate the value of entropy‐based measures, yet they operate on continuous representations rather than discrete relational graphs. Structural‐quality metrics derived from ontology hierarchy complexity (e.g., Inverse Multiple Inheritance) quantify schema simplicity~\cite{seo2022structuralqualitymetricsevaluate}, but overlook the rich spectral and semantic diversity inherent in multi‐relational KGs. Extensions of spectral analysis to temporal KGs via hyperbolic graph embeddings highlight how evolving relations can collapse spectral class separation and hinder tail prediction~\cite{10.1609/aaai.v38i11.29173}, yet they remain specialized to dynamic settings.

Spectral measures such as the Cumulative Spectral Gradient (CSG) have been proposed to capture dataset complexity by quantifying class separability via eigenvalue spectra of the normalized graph Laplacian~\cite{branchaud2019spectral}. In image classification, higher CSG correlates with lower test accuracy. However, KG link prediction poses a distinct challenge: each head–relation pair defines a multi‐class task over thousands of candidate tails, and embeddings may derive from transformers or translational models. Two central claims of CSG warrant re‐evaluation in this regime:
\begin{itemize}
  \item {\bf Scalability}: CSG’s dependence on the nearest‐neighbor parameter $K$ may not generalize when the number of classes scales to KG sizes.
  \item {\bf Predictive power}: It remains unclear whether CSG values computed over KGE embeddings correlate with standard link‐prediction metrics such as MRR and Hits@k.
\end{itemize}
To date, no work has systematically examined CSG’s sensitivity to its hyperparameters $(K)$ or its stability on canonical KG benchmarks (e.g., FB15k-237, WN18RR).

Beyond spectral approaches, KG complexity also arises from semantic and structural factors. Semantic complexity can be captured by metrics such as {\em Relation Entropy}, which measures global uncertainty in relation distributions, and {\em Node‐level Maximum Relation Diversity}, which quantifies local ambiguity for each entity. Structural complexity emerges via network topology—properties like {\em average degree}, {\em degree entropy}, {\em PageRank}, and {\em eigenvector centrality} influence information flow and may impact model performance.

In this paper, we present the first comprehensive evaluation of KG dataset complexity across three complementary dimensions—{\em spectral}, {\em semantic}, and {\em structural}—and relate these metrics to link‐prediction performance. Our study covers five widely used benchmarks (FB15k-237, WN18RR, CoDEx-L/M/S), three classes of KGE models (translational, bilinear, and transformer‐based), and examines:
\begin{itemize}
  \item The {\bf sensitivity} of CSG to its hyperparameters $(K)$ and its {\bf scalability} with the number of relation classes.
  \item The {\bf correlation} between CSG and standard performance metrics (MRR, Hits@1, Hits@10).
  \item The {\bf predictive power} of semantic metrics (Relation Entropy, Relation Type Cardinality, Node‐level Max Relation Diversity) and structural metrics (Average Degree, Degree Entropy, PageRank, Eigenvector Centrality) with respect to MRR and Hits@k.
\end{itemize}

Our key findings are:
\begin{enumerate}
  \item \textbf{CSG Fragility and Unreliability}: CSG varies dramatically with $K$ and $M$, does not scale naturally with class cardinality, and exhibits near‐zero Pearson correlation with MRR and Hits@1 across all datasets and models.
  \item \textbf{Semantic Complexity as a Better Indicator}: Higher Relation Entropy, Node‐level Max Relation Diversity, and larger Relation Type sets correlate inversely with MRR and Hits@1, accurately reflecting dataset hardness.
  \item \textbf{Structural Connectivity and Recall}: Metrics such as Average Degree and Degree Entropy correlate positively with MRR and Hits@1, while centrality measures (Degree Centrality, Eigenvector Centrality, PageRank) align with looser recall metrics like Hits@10.
\end{enumerate}

These results challenge the utility of spectral separability measures like CSG for KG link prediction and demonstrate that semantic and structural metrics offer more robust, model‐agnostic complexity estimates. We release our code and dataset complexity profiles to enable complexity‐aware model evaluation and KG curation.  

\paragraph{Contributions}
\begin{itemize}
  \item \textbf{First systematic evaluation} of CSG in large‐scale, multi‐class KG link‐prediction tasks, uncovering its sensitivity and lack of predictive power.
  \item \textbf{Unified complexity assessment} integrating spectral, semantic, and structural metrics tailored to the multi‐relational nature of KGs.
  \item \textbf{Empirical insights} showing that semantic ambiguity (Relation Entropy, Relation Diversity) and structural connectivity (Degree, Centrality) correlate more faithfully with MRR and Hits@k than CSG.
  \item \textbf{Benchmark complexity profiles} for FB15k-237, WN18RR, CoDEx-L/M/S, facilitating principled dataset selection and model diagnosis.
\end{itemize}

\section{Methodology}
To systematically evaluate knowledge graph complexity, we adopt a multi-faceted approach that examines spectral, semantic, and structural features across five standard benchmarks (FB15k-237, WN18RR, CoDEx-L/M/S). We first assess the sensitivity of the Cumulative Spectral Gradient (CSG) to its key parameter $K$ then analyze its correlation with model performance (MRR, Hits@k) to test its viability as a complexity metric. For semantic and structural characterization, we compute relation entropy, node-level diversity, degree-based metrics, and centrality measures, evaluating their predictive power through correlation analysis with link-prediction outcomes. This comprehensive framework enables us to identify robust, task-aligned indicators of KG complexity.

\textbf{Problem Statement:} The main objective of this study is to formulate and quantify the complexity of link/tail prediction in knowledge graphs (KGs) from both structural and spectral perspectives. We represent an KG dataset as a set of triplets \( T = \{(h, r, t) \mid h, t \in E, r \in R\} \),  where \( E \) denotes the set of entities and \( R \) the set of relations. Given a collection of KG datasets \( \mathcal{D} = \{D_1, D_2, \ldots, D_N\} \), each defined over its own entity and relation sets \( (E_i, R_i) \), evaluate a complexity measure function \( f: \mathcal{D} \rightarrow \mathbb{R} \) such that \( f(D_i) > f(D_j) \) implies that \( D_i \) is more complex than \( D_j \) with respect to the average link(tail) prediction performance across KGC models. Below, we explain the proposed methodology step by step. We first describe each of the metrics we calculate and begin with how we apply CSG metric to the link prediction task.

\subsection{Cumulative Spectral Gradient (CSG)}
\begin{figure}[ht]
\begin{center}
\centerline{\includegraphics[width=1.1\columnwidth]{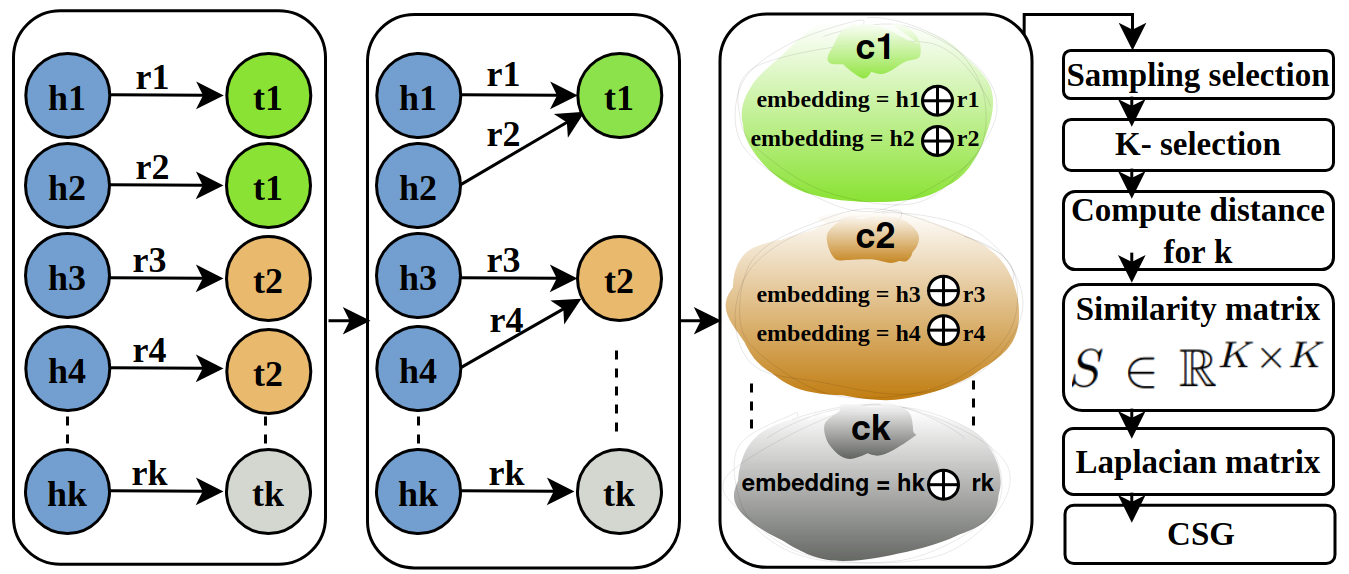}}
\caption{\scriptsize Left box shows triplets where the heads are $(h_1, h_2, \ldots, h_k)$ (green), relations $(r_1, r_2, \ldots, r_k)$, tails $(t_1, t_2, t_3)$ (blue, yellow, purple). The next box denotes the grouping of tail entities into classes: $C_1$ for $t_1$, with $(h_1, r_1, t_1)$ and $(h_2, r_2, t_1)$ belonging to the same class. BERT embeds head-relation pairs into 768-dimensional vectors, concatenated as $h_1 \oplus r_1$, $h_2 \oplus r_2$ for class $C_i$. A sampled $k$-nearest neighbor search computes distances and a similarity matrix $S \in \mathbb{R}^{K \times K}$. The Laplacian matrix $L$ is derived, and the CSG is calculated from its eigenvalue gaps, reflecting the spectral complexity of the KG.}
\label{m-csg}
\end{center}
\end{figure}

To compute the Cumulative Spectral Gradient (CSG), we transform KG triplets into multi-class representations, use BERT embeddings for semantic richness, and apply spectral analysis to derive CSG values, which quantify class overlap in the embedding space; see Figure \ref{m-csg} for details.

\textit{Grouping by Tail Entities:} Knowledge graphs, consist of a set of triplets:
\begin{equation}
T = \{(h_i, r_i, t_i) \mid h_i \in E, r_i \in R, t_i \in E \},
\end{equation}
where \( h_i \) is the head entity, \( r_i \) is the relation, and \( t_i \) is the tail entity, with \( E \) being the set of all entities and \( R \) the set of all relations. The next step organizes this data by grouping triplets according to their tail entities class \(C\) (each unique \( t_i \rightarrow \) denotes a unique class \( C_i \)) using a mapping function:
\begin{equation}
G(C_i) = \{ (h, r) \mid (h, r, C_i) \in T \}, \quad \forall C_i \in E,
\end{equation}
resulting in a mapping:
\begin{equation}
C_i \mapsto G(C_i),
\end{equation}
which aggregates all \( (h, r) \) pairs pointing to the same tail \( C_i \). Each unique tail entity is treated as a distinct class, forming a set:
\begin{equation}
C_i = \{C_1, C_2, \ldots, C_M\},
\end{equation}
where $ M = |\{t_i\}| $ is the total number of unique tail entities. This forms $ M $ classes for downstream analysis.

\textit{Generating Embeddings:}  
To transform textual head entities and relations into numerical form, a pre-trained BERT model generates dense vector embeddings. For each head entity \( h \) and relation \( r \), embeddings are:
\begin{equation}
e_h = \text{BERT}(h) \in \mathbb{R}^d, \quad e_r = \text{BERT}(r) \in \mathbb{R}^d,
\end{equation}
\( d \) the embedding dimension is. BERT-Base (Hugging Face Transformers) was used to generate 768-dimensional embeddings, preprocessing head entities and relations as single tokens. For every triplet \( (h, r, C_i) \in T \), a composite vector is formed:
\begin{equation}
\phi(h, r) = e_h \oplus e_r \in \mathbb{R}^{2d},
\end{equation}
where \( \oplus \) denotes concatenation. These composite vectors are then grouped according to their corresponding tail entities:
\begin{equation}
\Phi(C_i) = \{ \phi(h, r) \mid (h, r, C_i) \in T \},
\end{equation}
each tail \( C_i \) is associated with a set of \( (h, r) \) vectors.
This step provides a meaningful representation of the triplet data, organized by tail classes, preparing the data for complexity analysis.
\textit{Similarity Computation and Matrix Construction:}  
A similarity matrix $ S \in \mathbb{R}^{M \times M} $ is constructed, where \( M \) is the number of classes and $ S_{ij} $ measures how often samples from class $ C_i $ have neighbors in $ C_j $, indicating inter-class overlap. Each vector:
\begin{equation}
\phi_m = e_h \oplus e_r \in \mathbb{R}^{2d}.
\end{equation}
A subset is sampled:
\begin{equation}
\begin{split}
N_s & = \min(N, |\Phi(C_i)|), \\
\Phi(C_i)_{\text{sample}} &= \{\phi_1, \phi_2, \ldots, \phi_{N_s}\} \subset \Phi(C_i)
\end{split}
\end{equation}
where \( N \) is the number of vector samples per class. 
To manage computational complexity for large KGs, we sample $N_s = 120$ vectors per class.
For each \( \phi_m \in \Phi(C_i)_{\text{sample}} \), compute its \( k = 50 \)-nearest neighbors via L2 distance:
\begin{equation}
    \|\phi_m - \phi_n\|_2^2 = \sum_{l=1}^{2d} (\phi_{m,l} - \phi_{n,l})^2.
\end{equation}
it computes the Euclidean distance between two concatenated BERT embeddings, 
\( \boldsymbol{\phi}_m \) and \( \boldsymbol{\phi}_n \), where \( \boldsymbol{\phi}_m, \boldsymbol{\phi}_n \in \mathbb{R}^{2d} \) represent the combined head-relation embeddings of triplets, respectively. while \( \phi_{m,l} \) and \( \phi_{n,l} \) denote the \( l \)-th components of the respective vectors. This distance metric is used during \( k \) values neighbor (k-NN) search to measure nearest neighbor triplets grouped by tail entities, enabling the construction of the class similarity matrix \( S \), which can be defined as in Equation \ref{eq:sim}. The distance computation directly impacts the spectral analysis by indicating how tightly or loosely classes overlap, thereby influencing the Cumulative Spectral Gradient (CSG), a measure of dataset complexity derived from the eigenvalue gaps in the graph Laplacian.
\begin{equation}
\label{eq:sim}
    S_{ij} = \frac{1}{N_s k} \sum_{\phi_m \in \Phi(C_i)_{\text{sample}}} \sum_{\phi_n \in N_K(\phi_m)} \mathbb{I}[\phi_n \in \Phi(C_j)],
\end{equation}
where $ \mathcal{N}_k(\phi_m) $ is the set of $ k $ nearest neighbors of $ \phi_m $, and:
\begin{equation}
    \mathbb{I}[\phi_n \in \Phi(C_j)] = \begin{cases}
1, & \text{if } \phi_n \in \Phi(C_j), \\
0, & \text{otherwise}.
\end{cases}
\end{equation}
$\phi_n$ is an embedding vector, $\Phi(C_j)$ denotes the set of embeddings for class $C_j$, and $I$ is an indicator function returning $1$ if $\phi_n$ belongs to $C_j$. It is employed in the formation of the similarity matrix $S$ to enumerate the $K$-nearest neighbors of the class $C_i$ that belong to $C_j$, quantifying inter-class overlap for complexity analysis.

\textit{Graph Laplacian and Spectral Analysis:} 
Graph Laplacian captures the connectivity and clustering tendencies of the classes, rooted in graph theory and spectral analysis. The normalized Laplacian provides a standardized measure of how classes are interconnected, accounting for variations in their degrees of connection. The graph Laplacian captures class connectivity and clustering tendencies. The diagonal degree matrix \( D \in \mathbb{R}^{M \times M} \) can be defined as Equation \ref{eq:dii}, while the normalized Laplacian as Equation \ref{eq:l}.
\begin{equation}
\label{eq:dii}
    D_{ii} = \sum_{j=1}^{M} S_{ij}, \quad D_{ij} = 0 \text{ for } i \ne j.
\end{equation}
\begin{equation}
\label{eq:l}
    L = I - D^{-1/2} S D^{-1/2},
\end{equation}
where \( I \) is the \( M \times M \) identity matrix, and:
\begin{equation}
D^{-1/2}_{ii} = 
\begin{cases}
1/\sqrt{D_{ii}}, & D_{ii} > 0, \\
0, & \text{otherwise}.
\end{cases}
\end{equation}
$D_{ii} = \sum_{j=1}^{M} S_{ij}$ representing the total similarity of a class $C_i$ to all other classes, where $D_{ii}^{-1/2} = \frac{1}{\sqrt{D_{ii}}}$, ensures eigenvalues. Compute eigenvalues \( \lambda_0, \lambda_1, \ldots, \lambda_{K-1} \) and eigenvectors \( u_1, u_2, \ldots, u_M \) from:
\begin{equation}
\label{eq:lui}
    L u_i = \lambda_i u_i, \quad u_i \in \mathbb{R}^M, \quad \|u_i\| = 1, \quad 0 \le \lambda_i \le 2.
\end{equation}
ordered non-decreasingly: 
\begin{equation}
    0 = \lambda_0 \le \lambda_1 \le \ldots \le \lambda_{M-1},
\end{equation}
yields eigenvalues $\lambda_i$ and orthonormal eigenvectors $u_i$, which encode structural properties.

\textit{Cumulative Spectral Gradient (CSG) Computation:}  
CSG measures complexity by summing the differences between consecutive Laplacian eigenvalues, capturing how the graph’s structure evolves across its spectrum. It reflects class separation and global graph properties, offering insight into prediction difficulty—especially relevant for tail prediction. Order the eigenvalues:
\begin{equation}
    \text{Define gaps,} ~ \delta_i = \lambda_{i+1} - \lambda_i, \quad i = 0, 1, \ldots, M-2,
\end{equation}
\begin{equation}
    \text{Then,} ~\text{CSG}_{k_c} = \sum_{i=0}^{k_c - 1} \delta_i = \lambda_{k_c} - \lambda_0,
\end{equation}
\begin{equation}
   \text{and,} ~  \text{CSG}_{K-1} = \lambda_{K-1} - \lambda_0.
\end{equation}
The author of the study \cite{branchaud2019spectral} claims that higher CSG values indicate higher complexity (more class overlap); lower CSG values indicate better separation and easier tail prediction.

\textbf{Relation Types $ R_{\text{count}} $ (Combinatorial Complexity):}  The number of unique type relations $ R_{\text{count}} = |R| $ reflects the combinatorial richness of a KG’s schema and has a direct impact on the search space of link prediction models:
\begin{equation}
R_{count} = |R|
\end{equation}

\begin{equation}
\bm{R}_{\text{count}} = |\bm{R}|
\end{equation}

A larger number of relation types increases the dimensionality and expressiveness of the relational model, which directly affects the computational and statistical difficulty of learning effective embeddings. Moreover, in rule-based reasoning systems, the presence of more relations enables richer logical expressions and inference paths, further amplifying complexity.

\textbf{Relation Entropy \( H(R) \) (Information-Theoretic Uncertainty):} This metric quantifies the unpredictability in relation distribution, capturing how evenly relations are used across the graph:
\begin{equation}
p(r_k) = \frac{|\{ (h, r_k, t) \in T \}|}{|T|}
\end{equation}
\begin{equation}
H(R) = -\sum_{r_k \in R} p(r_k) \log_2 p(r_k)
\end{equation}
High entropy implies uniform usage and harder predictability. In the context of KGs, high entropy makes it harder for models to exploit frequency patterns during training. Relational entropy is rooted in  Shannon’s information theory, where higher entropy indicates greater uncertainty.  From a probabilistic modelling perspective, this translates into increased model calibration difficulty, since uniform distributions require more nuanced confidence estimation than skewed ones. Furthermore, entropy is linked to the concept of information gain in reasoning systems. Low entropy suggests dominant rules exist, while high entropy implies more uncertain relationships.

\textbf{Maximum Relation Diversity (Local Ambiguity):} This metric identifies local complexity hotspots by measuring the number of distinct relations connected to any single entity:
\begin{equation}
\text{Div}(e) = |\{ r \mid (e, r, \cdot) \in T \lor (\cdot, r, e) \in T \}|, \quad \forall e \in E
\end{equation}
\begin{equation}
\text{MaxRelDiv} = \max_{e \in E} \text{Div}(e)
\end{equation}
It captures entity-level heterogeneity, which is critical in real-world KGs where some nodes act as hubs connecting diverse types of relationships. From a network science perspective, such hubs increase structural irregularity, which has been shown to degrade performance in embedding models due to conflicting neighborhood signals. In particular, entities with high \( \text{Div}(e) \) are likely to appear in multiple contexts, leading to semantic ambiguity and embedding distortion in vector spaces. This phenomenon is closely related to the challenge of polysemy in natural language, where a single symbol (e.g., word or entity) represents multiple meanings.

In addition to the metrics described above (CSG, relation types, relation entropy, and maximum relation diversity), we computed several other structural and spectral measures to comprehensively characterize KG complexity. These include graph-theoretic metrics like modularity, betweenness centrality, and algebraic connectivity, as well as statistical properties such as average degree and clustering coefficients. All these properties are defined in the supplementary materials. A complete listing of these measures across all evaluated datasets is provided in Table~\ref{tab:1}, which shows their values for standard benchmarks including FB15k-237, CoDEx variants, and WN18RR. This multi-faceted analysis enables us to correlate different aspects of graph complexity with model performance in subsequent sections.

\subsection{Experiments}
This study utilized the following KG datasets for the experiments.
\begin{itemize}
    \item \textbf{FB15k-237} \citep{bollacker2008freebase}: An updated version of FB15k with inverse triplets removed to increase difficulty. It consists of 14,541 entities, 237 relations, 272,115 training, 17,535 validation, and 20,466 test triplets.
    \item \textbf{WN18RR} \cite{miller1995wordnet}: A subset of WN18, where reverse triplets are removed for increased complexity. The dataset includes 40,943 entities, 11 relations, 86,835 training, 2,924 validation, and 2,824 test triplets.
    \item \textbf{CoDEx-S} \cite{safavi2020codex}: A multi-domain KG sourced from Wikidata, featuring 2,034 entities, 42 relations, 32,888 training, 3,654 validation, and 3,656 test triplets.
    \item \textbf{CoDEx-M} \cite{safavi2020codex}: A medium-sized version of CoDEx KG with 17,050 entities, 51 relations, and 185,584 triples, designed for complex KGC tasks such as link prediction and triple classification.
    \item \textbf{CoDEx-L} \cite{safavi2020codex}: It includes 77,951 entities, 69 relations, and a total of 673,872 triplets.
\end{itemize}
\section{Results}
We first analyze the behavior of the Cumulative Spectral Gradient (CSG) as a complexity measure, examining its sensitivity and predictive power. Following this, we evaluate semantic and structural features to identify robust indicators of knowledge graph complexity, revealing how different feature types distinctly impact prediction performance.

\begin{figure}[ht]
\begin{center}
\centerline{\includegraphics[width=1.0 \columnwidth]{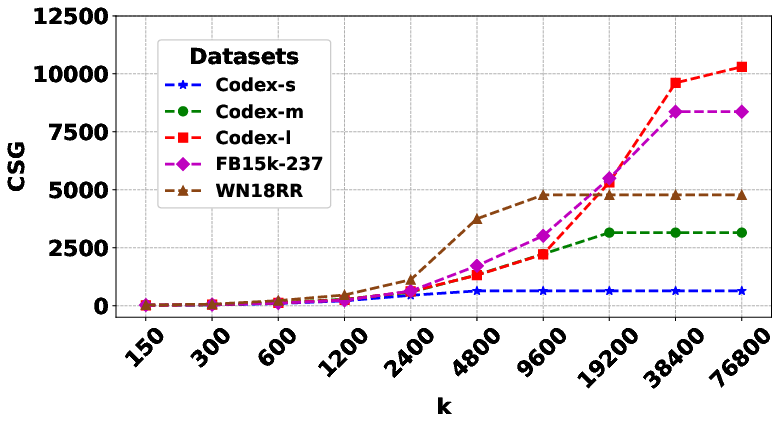}} 
\caption{\scriptsize A plot of CSG as a function of $K$ values at $M = 100$. $M$ is the number of Samples. }
\label{k_csg}
\end{center}
\end{figure}

\begin{figure}[ht]
\begin{center}
\centerline{\includegraphics[width=1.1\columnwidth]{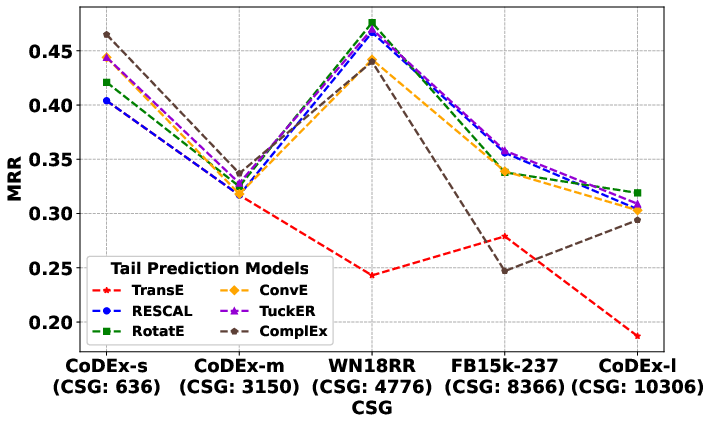}} 
\caption{\scriptsize Relationship Between MRR from different tail-prediction models on five standard KG datasets and the corresponding CSG values.}
\label{k-s-csg-emb}
\end{center}
\end{figure}

\textbf{Sensitivity and Predictive Power of CSG:} Figure \ref{k_csg} illustrates how the CSG is strongly influenced by the parameter $K$, with CSG values increasing consistently as $K$ increases across all datasets. This trend reveals that larger $K$-values capture broader structural patterns, leading to higher perceived complexity, while smaller $K$-values emphasize local structure and result in lower CSG. Furthermore, the variation in CSG across different datasets highlights the importance of tailoring $K$ to the specific structural and semantic characteristics of each KG. Figure~\ref{k-s-csg-emb} plots CSG values for five standard KG benchmarks against the corresponding MRR values achieved by a suite of tail‐prediction models. Contrary to \citet{branchaud2019spectral}, we observe no meaningful correlation (mean Pearson coefficient \(R = - 0.644\)) between CSG and model performance across all datasets and methods. In summary, contrary to the assertion that CSG can consistently forecast downstream performance, these findings cast doubt on its value as a model‐independent separability measure for large‐scale classification tasks specially in KG domain and underscore the necessity for more reliable metrics in KG evaluation. 

\begin{table*}[t]
\centering
\caption{Dataset Statistics and Measures}
\label{tab:1}
\resizebox{\textwidth}{!}{%
\renewcommand{\arraystretch}{1.2} 
\begin{tabular}{lccccc}

\toprule
\textbf{Metric} & \textbf{FB15k-237} & \textbf{CoDEx-L} & \textbf{CoDEx-M} & \textbf{CoDEx-S} & \textbf{WN18RR} \\
\midrule
CSG &8366&10306&3150&636&4776\\
Relation Entropy & 6.5527 & 3.9945 & 3.8017 & 3.4038 & 2.2323 \\
Relation Types & 237.0000 & 69.0000 & 51.0000 & 42.0000 & 11.0000 \\
Max Relation Diversity & 55.0000 & 21.0000 & 19.0000 & 13.0000 & 7.0000 \\
Edge Betweenness Centrality (mean) & 0.0001 & 0.0000 & 0.0000 & 0.0000 & 0.0000 \\
Modularity & 0.7986 & 0.4378 & 0.4289 & 0.5220 & 0.4486 \\
Structural Entropy (Community-based) & 4.3930 & 2.0152 & 1.7319 & 1.4244 & 2.2392 \\
Homophily (Degree Assortativity) & -0.0547 & -0.1057 & -0.1250 & -0.2273 & -0.1359 \\
Average Degree & 4.5379 & 15.4272 & 23.8097 & 35.5978 & 39.6353 \\
Degree Centrality (mean) & 0.0001 & 0.0002 & 0.0014 & 0.0175 & 0.0028 \\
Betweenness Centrality (mean) & 0.0001 & 0.0000 & 0.0000 & 0.0001 & 0.0002 \\
Closeness Centrality (mean) & 0.0297 & 0.0061 & 0.0092 & 0.0633 & 0.2214 \\
Eigenvector Centrality (mean) & 0.0003 & 0.0002 & 0.0008 & 0.0063 & 0.0018 \\
PageRank (mean) & 0.0000 & 0.0000 & 0.0001 & 0.0005 & 0.0001 \\
Local Clustering Coefficient (mean) & 0.0657 & 0.0980 & 0.0608 & 0.0952 & 0.2040 \\
Global Clustering Coefficient & 0.0657 & 0.0980 & 0.0608 & 0.0952 & 0.2040 \\
Transitivity & 0.0097 & 0.0008 & 0.0034 & 0.0594 & 0.0172 \\
Algebraic Connectivity & -0.0000 & 0.0000 & 0.8583 & 3.5251 & -0.0000 \\
Spectral Gap & 0.2046 & 106.4123 & 80.0789 & 79.6931 & 6.2834 \\
Degree Entropy & 3.2645 & 3.9721 & 4.1978 & 4.9232 & 6.4799 \\
Chromatic Number & 5.0000 & 25.0000 & 24.0000 & 23.0000 & 29.0000 \\
Girth & 1.0000 & 3.0000 & 3.0000 & 3.0000 & 1.0000 \\
\bottomrule
\end{tabular}%
}
\end{table*}

\begin{figure}[h!]
\begin{center}
\centerline{\includegraphics[width=1\columnwidth]{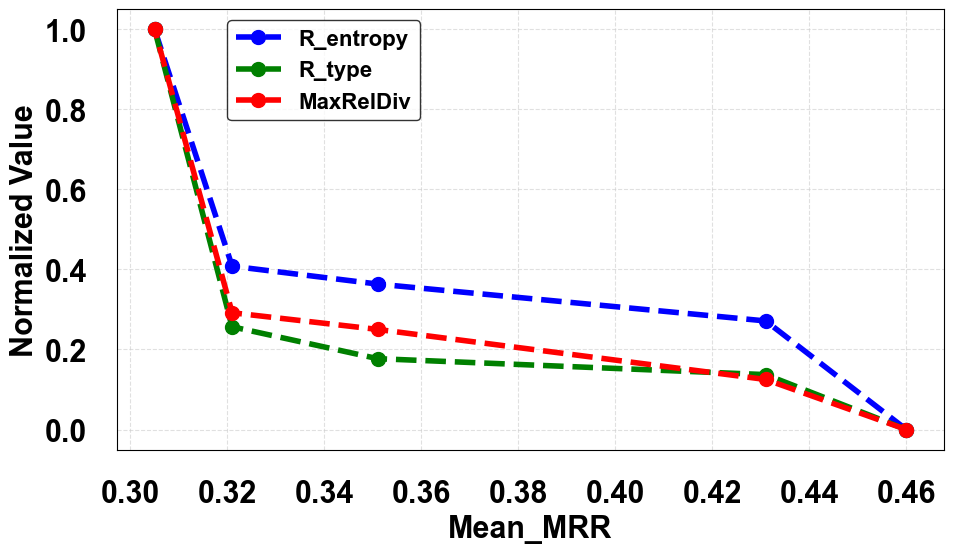}}
\caption{\scriptsize ean Mean MRR vs. Normalized Features — Higher feature values correspond to higher Mean MRR, indicating easier link prediction.}
\label{mrr_f1}
\end{center}
\end{figure}

\begin{figure}[h!]
\begin{center}
\centerline{\includegraphics[width=1\columnwidth]{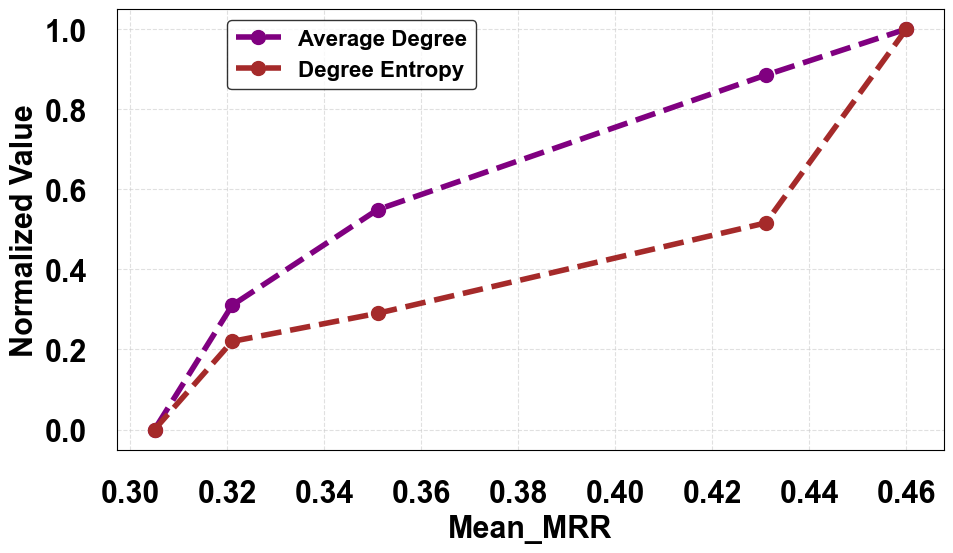}}
\caption{\scriptsize Mean MRR vs. Normalized Features — Higher feature values correspond to higher Mean MRR, indicating easier link prediction.}
\label{mrr_f2}
\end{center}
\end{figure}

Figures \ref{mrr_f1} and \ref{mrr_f2} reveal distinct relationships between KG complexity features and link-prediction performance. Figure \ref{mrr_f1} shows that higher values of semantic features—relation entropy, relation type cardinality, and node-level relation diversity—inversely correlate with improved Mean MRR, suggesting thathigher these relatoin properties lower the predictions performance. In contrast, Figure \ref{mrr_f2} demonstrates that structural features like average degree and degree entropy are positively influence Mean MRR, indicating that densely connected graphs facilitate easier link prediction.

\begin{figure}[h!]
\begin{center}
\centerline{\includegraphics[width=1\columnwidth]{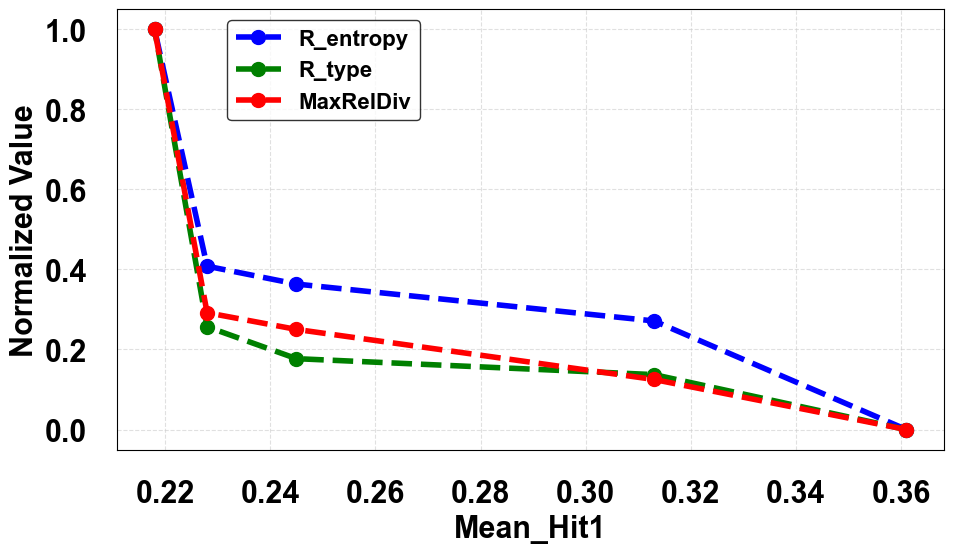}}
\caption{\scriptsize Mean Hit@1 vs. Normalized Features — Higher feature values correspond to higher Mean Hit@1, indicating easier link prediction.}
\label{hit1_f1}
\end{center}
\end{figure}

\begin{figure}[h!]
\begin{center}
\centerline{\includegraphics[width=1\columnwidth]{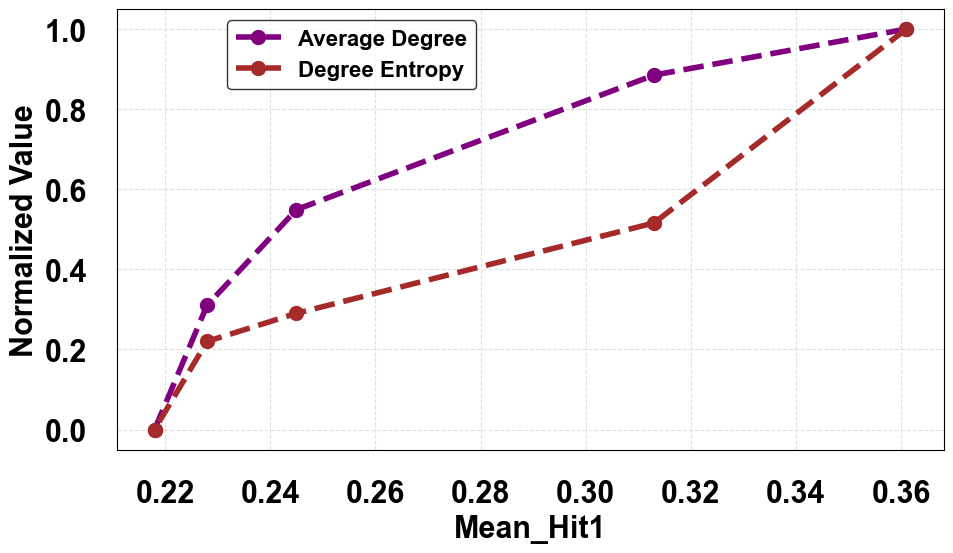}}
\caption{\scriptsize Mean Hit1 vs Normalized Features}
\label{hit1_f2}
\end{center}
\end{figure}

Figures \ref{hit1_f1} and \ref{hit1_f2} demonstrate how different knowledge graph features influence prediction difficulty. Semantic features like relation entropy, relation type diversity, and node-level relation complexity show an inverse relationship with performance - when these features increase, prediction accuracy (Mean Hit@1) decreases, indicating they represent challenging aspects of knowledge graphs that make link prediction harder. Conversely, structural features including average node degree and degree entropy exhibit a positive correlation with performance - higher values of these connectivity metrics correspond to better prediction results, showing they provide beneficial structural patterns that models can effectively learn from. These trends highlight that features with higher values linked to lower MRR increase complexity, while those with higher values tied to higher MRR aid prediction simplicity.

\begin{figure}[h!]
\begin{center}
\centerline{\includegraphics[width=1\columnwidth]{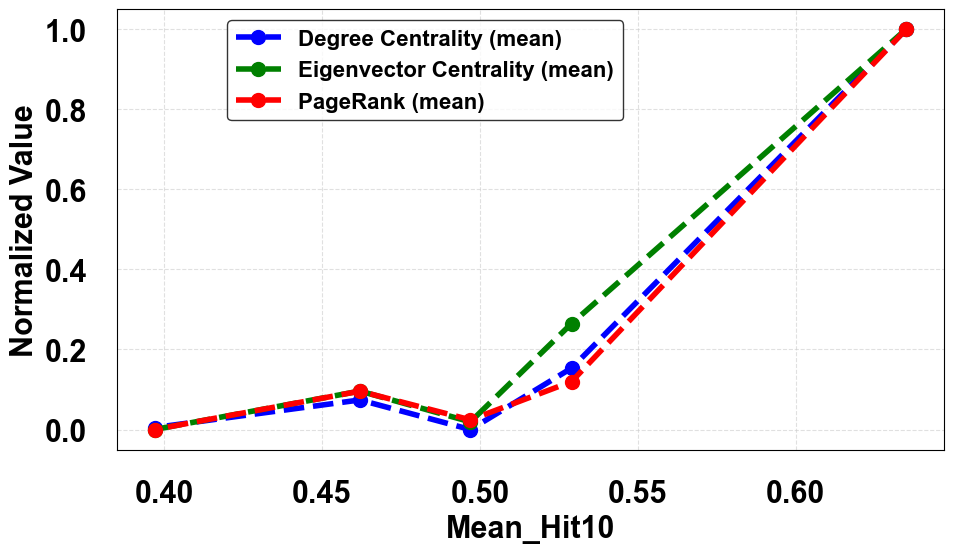}}
\caption{\scriptsize Mean Hit10 vs Normalized Features}
\label{hit10_f1}
\end{center}
\end{figure}

The figure \ref{hit10_f1} demonstrates how centrality metrics influence link prediction performance at the Hits@10 level. As shown in the plot, higher values of degree centrality, eigenvector centrality, and PageRank correlate strongly with improved Mean Hit10 scores, indicating these structural features significantly enhance prediction performance. This positive relationship occurs because nodes with greater centrality (more connections, higher influence, or greater importance in the network) provide richer structural patterns that models can effectively utilize for tail entity prediction. The consistent upward trend across all three centrality measures confirms that well-connected, influential nodes in knowledge graphs serve as reliable anchors that make top-10 predictions more accurate, highlighting how structural prominence facilitates rather than hinders link prediction tasks.

\paragraph{Structural Complexity and Link‐Prediction Performance:}  
Our empirical analysis reveals a clear dichotomy between semantic heterogeneity and structural connectivity in driving link‐prediction metrics. Measures of {\em semantic complexity}—Relation Entropy, Node‐level Maximum Relation Diversity, and Relation Type Cardinality—show strong negative correlations with MRR and Hit@1, indicating that a proliferation of relation types and unpredictable relational distributions likely fragments the model embedding space and impedes fine‐grained discrimination at the top rank. In contrast, {\em local connectivity} metrics—Average Degree and Degree Entropy—correlate positively with MRR and Hit@1, suggesting that denser, variably‐connected neighbourhoods provide redundant multi‐hop cues that help models confidently surface the correct tail. Finally, {\em global prominence} metrics (Degree Centrality, Eigenvector Centrality, PageRank) align closely with Hit@10, reflecting that looser recall tolerates semantic ambiguity in favour of nodes with high overall visibility. Altogether, these findings expose a precision–recall trade‐off: high semantic diversity degrades top‐rank accuracy, whereas structural density boosts it, and global centrality governs broader candidate coverage.

\paragraph{Implications for Model Selection and Dataset Curation}  
These structural–semantic correlations offer concrete guidance for downstream deployment and fair benchmarking. For high‐precision tasks—such as pinpointing specific drug–target interactions—one should favor models with strong local discrimination (e.g., margin‐ranking losses paired with dense‐graph inductive biases) and curate evaluation sets with low relation entropy and moderate relation‐diversity. Conversely, when broad recall is paramount (e.g., recommendation or exploratory QA), emphasizing global‐centrality features and measuring Hits@10 will reward models adept at surfacing prominent but semantically diverse candidates. More generally, by profiling datasets along our proposed metrics, practitioners can anticipate generalization gaps—recognizing that high semantic heterogeneity intrinsically raises task difficulty—and compare methods on a complexity‐aware basis, ensuring that model improvements reflect algorithmic advances rather than dataset idiosyncrasies.

\section{Conclusion}
We presented the first systematic evaluation of Cumulative Spectral Gradient in KG link prediction and exposed its sensitivity and lack of correlation with standard performance metrics. By integrating spectral, semantic, and structural measures, we demonstrated that properties like Relation Entropy and Relation Diversity inhibit top‐rank accuracy, while graph density and centrality govern both precision and recall. Our complexity framework enables practitioners to anticipate generalization performance, compare models on a complexity‐aware basis, and curate datasets tailored to downstream objectives.

\bigskip
\noindent Thank you for reading these instructions carefully. We look forward to receiving your electronic files!

\bibliography{aaai2026}

\end{document}